\documentclass[10pt,a4paper]{article}

\usepackage{cogsci}

\cogscifinalcopy 
\usepackage{amsmath,amssymb,amsthm,amsfonts,graphicx,hyperref,rotating,epsfig,multirow,makecell,booktabs,stfloats,subfig,url}

\usepackage{pslatex}
\usepackage{apacite}
\usepackage{float} 
\usepackage{balance}

\title{InferEM: Inferring the Speaker's Intention for Empathetic Dialogue Generation}
 
\author{{\large \bf Guoqing Lv$^{1}$, Jiang Li \thanks{Guoqing Lv and Jiang Li contributed equally to this work.}$^{,2}$, Xiaoping Wang \thanks{Corresponding authors: Jiang Li and Xiaoping Wang.}, Zhigang Zeng} \\
The School of Artificial Intelligence and Automation, and the Key\\ Laboratory of Image Processing and Intelligent Control of Education Ministry of China, and the Hubei Key\\ Laboratory of Brain-inspired Intelligent Systems, Huazhong University of Science and Technology, Wuhan 430074, China\\
\{guoqinglv, lijfrank, wangxiaoping, zgzeng\}@hust.edu.cn\\
Accepted by the 45th Annual Meeting of the Cognitive Science Society (CogSci 2023)
}

\begin{document}

\maketitle

\begin{abstract}
Current approaches to empathetic response generation typically encode the entire dialogue history directly and put the output into a decoder to generate friendly feedback. These methods focus on modelling contextual information but neglect capturing the direct intention of the speaker. We argue that the last utterance in the dialogue empirically conveys the intention of the speaker. Consequently, we propose a novel model named InferEM for empathetic response generation. We separately encode the last utterance and fuse it with the entire dialogue through the multi-head attention based intention fusion module to capture the speaker's intention. Besides, we utilize previous utterances to predict the last utterance, which simulates human's psychology to guess what the interlocutor may speak in advance. To balance the optimizing rates of the utterance prediction and response generation, a multi-task learning strategy is designed for InferEM. Experimental results demonstrate the plausibility and validity of InferEM in improving empathetic expression.

\textbf{Keywords:} 
empathetic response generation, multi-head attention, multi-task learning
\end{abstract}

\section{Introduction}
\label{sec:intro}
Empathetic emotion, a reaction of one individual to the observed experiences of the interlocutor~\cite{davis1983measuring}, is a fundamental trait of human beings. Empathy is a contentious and complex emotion that emerges when we respond to the needs or suffering from another. Empathy has been applied in a wide range of domains, such as volunteering~\cite{batson1997empathy}, psychotherapy~\cite{bohart1997empathy}, charitable giving~\cite{pavey2012help} and longevity~\cite{poulin2013giving}. Equipping AI systems with the ability to express empathy can improve the human-machine interaction in many fields such as mental health support~\cite{sharma2023human} and social chatbot~\cite{zhou2020design}.

Empathetic dialogue systems are designed to comprehend the user's emotions and generate humanized responses, which facilitate enabling a better user experience~\cite{liu2021towards}. There have been a great deal of fruitful studies~\cite{hu2021mmgcn,li2022graphcfc} on emotion recognition. Recently, some attempts have been made to improve the quality of the generated empathetic response. MoEL~\cite{lin2019moel} computes the emotion distribution of the dialogue history and softly fuses the outputs of decoders, which is more interpretable than previous models and can select the correct decoder to effectively generate empathetic responses. MIME~\cite{majumder2020mime} splits 32 emotions into positivity and negativity groups, and mimics the speaker's emotion by appropriately mixing emotions from distinct groups.~\citeA{gao2021improving} design an emotion reasoner to judge context emotion labels and emotion cause-oriented labels, which are employed through gated attention networks to enhance the generation of empathetic responses. EmpHi~\cite{chen2022emphi} generates human-like responses with empathetic intents by means of computing the empathetic intent distribution of responses. KEMP~\cite{li2022knowledge} and CEM~\cite{sabour2022cem} leverage commonsense knowledge to enrich the dialogue history and generate more empathetic and informative responses. These approaches typically concatenate on encoding the representations of the entire dialogue history and recognizing the user's emotion but don't attach enough importance to capturing the direct intention of the speaker to generate a more targeted response. 

\begin{figure}[htbp]
	\centering
	\includegraphics[width=2.7in]{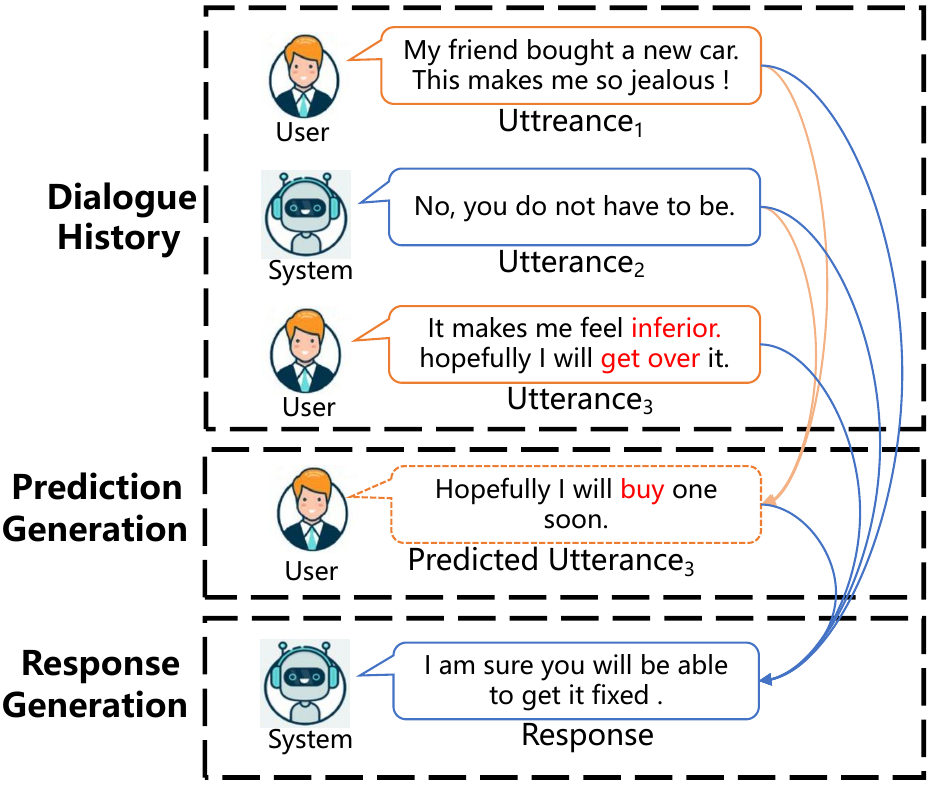}
	\caption{An example to illustrate how InferEM generates an empathetic response based on the dialogue history.}
	\label{fig:figure0}
\end{figure}

\citeA{zhang2018modeling} point out that the last utterance in a dialogue plays a crucial function, which empirically conveys the intention of interlocutor while previous utterances describe the dialogue in different aspects.~\citeA{garrod2009joint} assume that people can predict the interlocutor's response to enhance mutual understanding and promote successful communication. Inspired by their opinions, we propose a novel model to Infer the speaker's intention for EMpathetic dialogue generation (InferEM). The proposed model encodes separately the last utterance in the dialogue to capture the speaker's intention, then the multi-head attention~\cite{vaswani2017attention} is utilized to fuse the entire dialogue and the last utterance. Besides, InferEM leverages previous utterances to predict a virtual last utterance of the speaker, which is combined with the real one to improve the diversity of the speaker's intention. We design a multi-task learning strategy for InferEM to better optimize the parameters of the utterance prediction module and the response generation module. Figure~\ref{fig:figure0} is an example to illustrate how InferEM generates an empathetic response to the user based on the given dialogue history. The orange lines mean that the predicted $\mathrm{Utterance}_3$ is generated based on $\mathrm{Utterance}_1$ and $\mathrm{Utterance}_2$. The blue lines mean that the $\mathrm{Response}$ is generated based on all utterances in dialogue history and the predicted $\mathrm{Utterance}_3$. The red words imply the current intention of the user, so we separately encode the utterances containing these elements and offer them special attention.

Our contributions are as follows: $(1)$ We encode simultaneously the entire dialogue and the last utterance, then the multi-head attention based intention fusion network is adopted to integrate them. This process leads to a better understanding of the speaker's intention, which allows the model to generate more reliable empathetic responses. $(2)$ Aimed at promoting the diversity of the speaker's intention and generated response, we add a prediction module that simulates humans guessing the speaker's intention in advance based on the dialogue history. To the best of our knowledge, InferEM is the first attempt to leverage the predictive function to improve empathetic dialogue generation. $(3)$ We tailor a multi-task learning strategy for InferEM to balance the optimizing rates of the prediction module and the response module. Extensive experiments on a benchmark dataset verify the effectiveness of our proposed method.

\section{Preliminaries}
\subsection{Task Definition}
The dialogue history can be denoted as $C =  \left[U_1,U_2,\cdots,U_n \right]$, where each $U_i$ is the $i$-th utterance and $U_i$ includes $m_i$ words, denoted as $U_i = \left[w_i^1,w_i^2,\cdots,w_i^{m_i}\right]$. $[U_1,U_2,\cdots,U_n]$ may be uttered by the speaker or the listener. The target task is to generate an empathetic response based on dialogue history $C$. There is a preparatory task in our model that generate a prediction of $U_n$ according to the $n-1$ past utterances $C' =  \left[U_1,U_2,\cdots,U_{n-1} \right]$.

\subsection{Partial Notation}
\begin{figure}[ht]
  \centering
  \subfloat[EcEnc]{\includegraphics[width=1.15in]{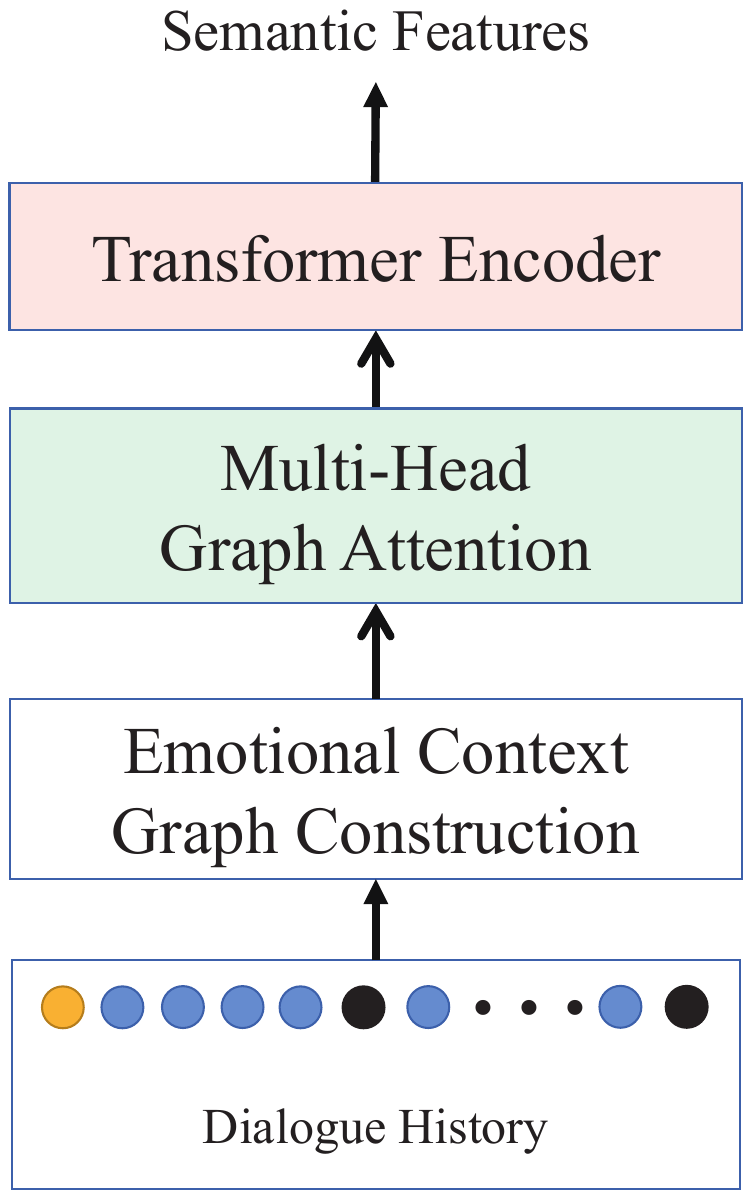}%
  \label{fig:figure1a}}
  \hfil
  \subfloat[Dec]{\includegraphics[width=2.05in]{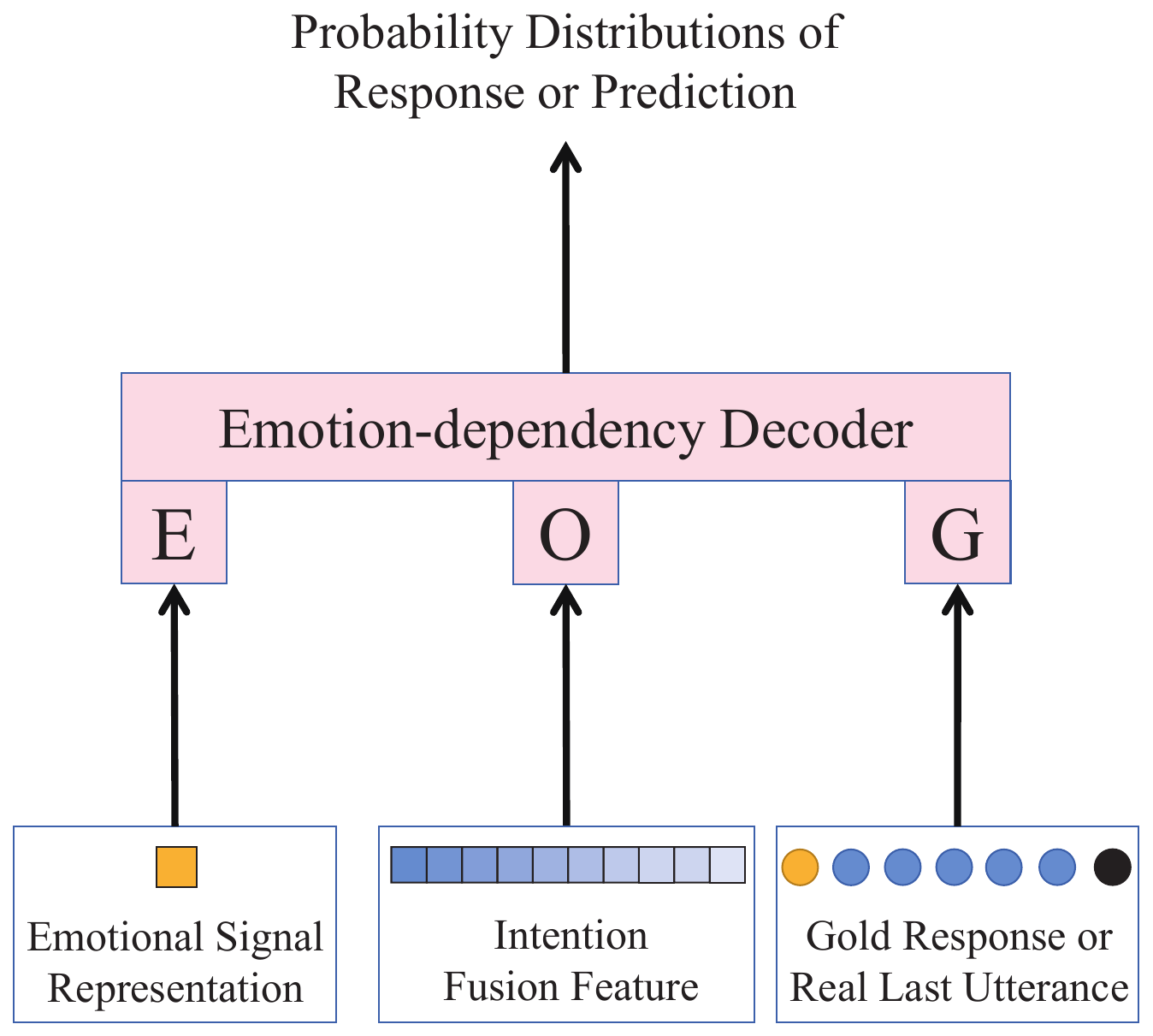}%
  \label{fig:figure1b}}
  \caption{The illustration of partial components.}
  \label{fig:figure1}
\end{figure}

As external knowledge can promote emotional understanding and expression, InferEM is furnished with the same external knowledge as KEMP~\cite{li2022knowledge}: $\left(1\right)$ we introduce emotion-related concepts~\cite{speer2017conceptnet} by constructing emotional context graph, and leverage both graph attention~\cite{velivckovic2017graph} and Transformer encoder~\cite{vaswani2017attention} to extract contextual semantic features. This process is called Emotion context encoder according to KEMP (see Figure~\ref{fig:figure1a}), which we simply denote as $\mathbf{EcEnc}$; $\left(2\right)$ we leverage emotion intensity values $\eta$ provided by NRC\_VAD~\cite{zhong2019knowledge} to calculate the emotion signal representation. 

Referring to KEMP, the Emotion-dependency decoder (we simply denote as $\mathbf{Dec}\left(E, O, G\right)$) is employed to generate the probability distributions of prediction or response, where $E$ indicates the emotion signal representation, $O$ is the semantic feature encoded by the Emotional context encoder, and $G$ indicates the gold prediction or response during training and the decoded words during testing. The Emotion-dependency decoder is shown in Figure~\ref{fig:figure1b}.

\begin{figure}[htbp]
	\centering
	\includegraphics[width=1.8in]{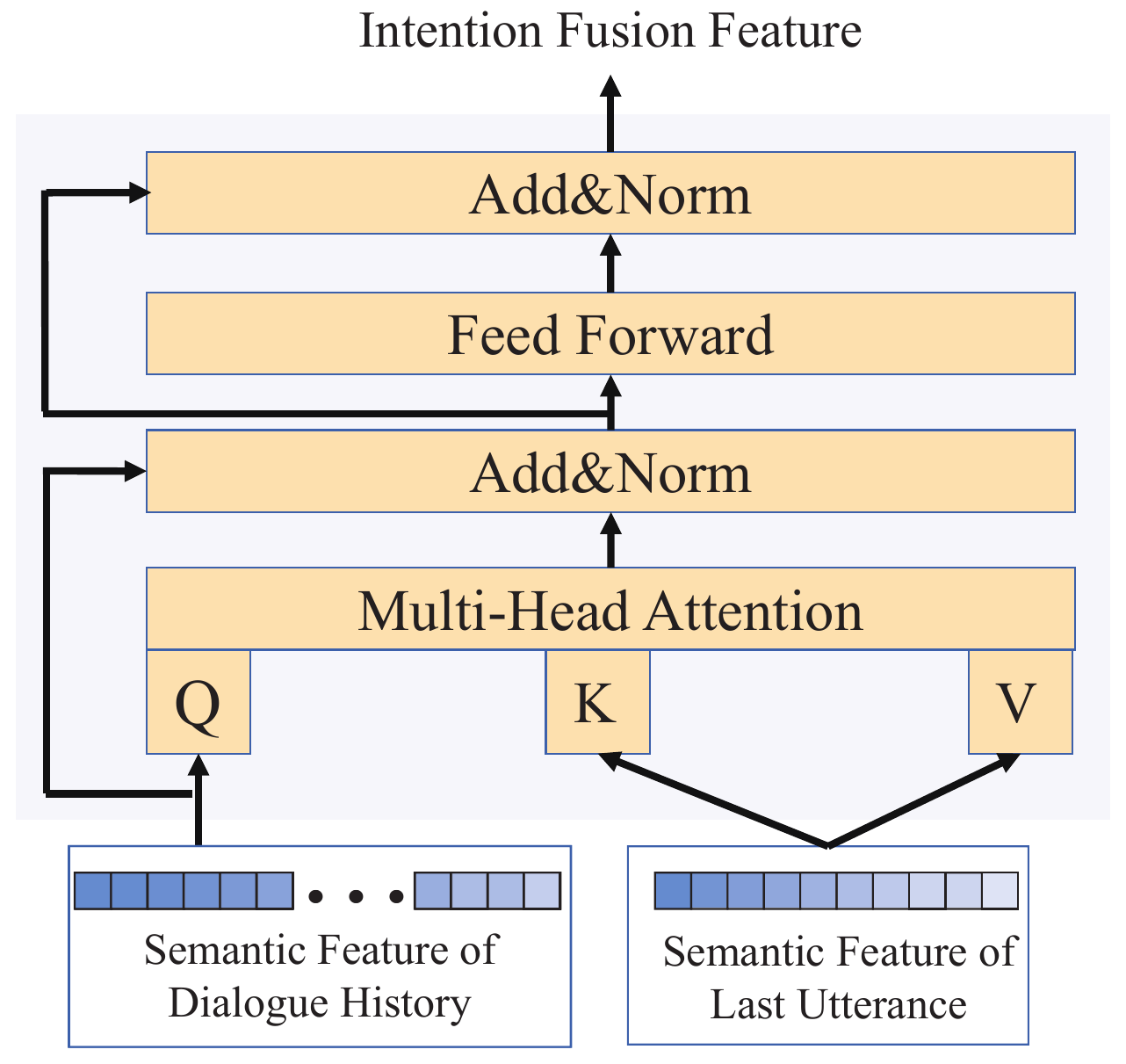}
	\caption{Multi-head attention based intent fusion network.}
	\label{fig:figure1c}
  \end{figure}
In our model, we adopt a Multi-head Attention based Intent Fusion Network (MAIFNet) by drawing on the structure of Transformer~\cite{vaswani2017attention}. MAIFNet mainly consists of a multi-head attention layer, a feedforward layer, two residual connections and two normalization layers, and the structure of MAIFNet is shown in Figure~\ref{fig:figure1c}. MAIFNet are utilized to fuse the entire dialogue history and the last utterance in the dialogue, which can perceive the intention of the person uttering the last utterance. We denote the multi-head attention as $\mathbf{Att}\left(Q, K, V\right)$, where $Q$, $K$ and $V$ indicate respectively the representations of query, key and value. The feedforward layer consists of two fully connected layers and is indicated as $\mathbf{FF}\left(X\right)$, where $X$ denotes the input of the feedforward layer.

Following previous work~\cite{wolf2019transfertransfo}, the embedding of the given dialogue history $C$ is computed as:
\begin{equation}
E_C = E_w(C) + E_p(C) + E_s(C), 
\end{equation}
where $E_w(C)$, $E_p(C)$ and $E_s(C)$ denote the word embedding, positional embedding and state embedding of dialogue history $C$.

\section{Methodology}
\begin{figure}[tbp]
  \centering
  \includegraphics[width=3.3in]{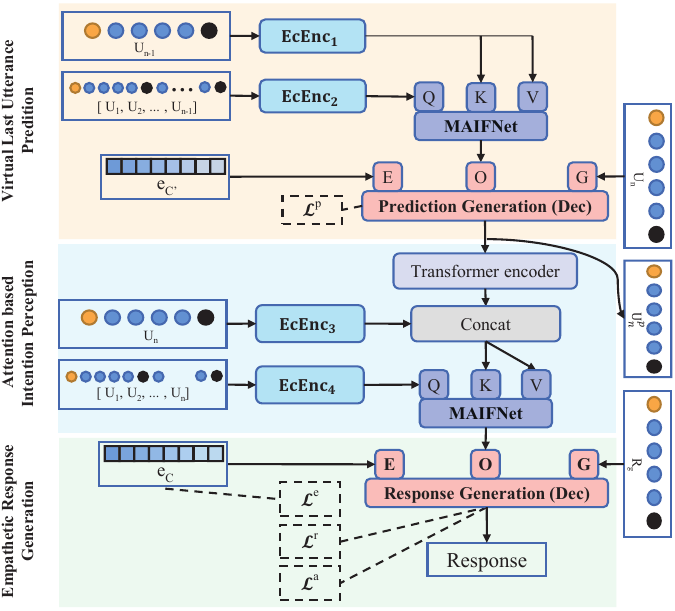}
  \caption{The architecture of InferEM.}
  \label{fig:figure2}
\end{figure}

The proposed InferEM mainly includes virtual last utterance prediction, attention based intention perception and empathetic dialogue generation, as shown in Figure~\ref{fig:figure2}. In the stage of last utterance prediction, the semantic features of $C'$ (the $n-1$ past utterances) and $U_{n-1}$ (the $(n-1)$-th utterance) are extracted by $\mathbf{EcEnc}$; on this basis, we fuse them by MAIFNet and then predict a virtual last utterance $U_n^p$. In the stage of intention perception, we first extract the semantic features of $C$ (the $n$ past utterances) and $U_n$ (the $n$-th utterance) by $\mathbf{EcEnc}$; then the semantic feature of $U_n$ and that of the virtual one are concatenated as that of the final last utterance $U_n^{pr}$ to enhance the diversity of the speaker's intention; finally, the semantic feature of $U_n^{pr}$ and that of the entire dialogue are fused by MAIFNet to perceive the intention of the speaker uttering the last utterance. And in the stage of empathetic dialogue generation, we feed the above result into $\mathbf{Dec}$ to generate an empathetic response.

\subsection{Virtual Last Utterance Prediction}\label{sec:last}
To extract the semantic feature of the $n-1$ past utterances $C'$ and that of the $(n-1)$-th utterance $U_{n-1}$, we first obtain their embeddings (i.e., $E_{C'}$ and $E_{U_{n-1}}$) and furnish them with emotion-related concepts according to previous work~\cite{li2022knowledge}; then $E_{C'}$ and $E_{U_{n-1}}$ are fed to $\mathbf{EcEnc}$:
\begin{equation}
  \begin{aligned}
    &S_{C'} = \mathbf{EcEnc}_1(E_{C'}),\\
    &S_{U_{n-1}} = \mathbf{EcEnc}_2(E_{U_{n-1}}),
  \end{aligned}
\end{equation}
where $E_{C'} \in \mathbb{R}^{q_{C'} \times f}$, $q_{C'}$ is the number of words in $C'$, $q_{C'}=m_1+\cdots+m_{n-1}$, and $f$ is the dimension of the word embedding; $E_{U_{n-1}} \in \mathbb{R}^{m_{n-1} \times f}$, $m_{n-1}$ is the number of words in $U_{n-1}$; $\mathbf{EcEnc}_1$ and $\mathbf{EcEnc}_2$ are Emotion context encoders with different emotion-related concepts and parameters.

As the virtual last utterance $U_{n}^{p}$ is predicted after the $\left(n-1\right)$-th utterance $U_{n-1}$ is spoken, we argue that the generated process of $U_{n}^{p}$ should adequately capture the intention of $U_{n-1}$. To this end, we utilize MAIFNet to fuse $S_{C'}$ and $S_{U_{n-1}}$, which is shown in the upper part of Figure~\ref{fig:figure2}. The fusion process can be formulated as: $F_{C'}' = \mathbf{MAIFNet}(S_{C'}, S_{U_{n-1}})$. Concretely, $\mathbf{MAIFNet}$ mainly includes a multi-head attention layer, a feed-forward layer, two residual connections and two normalization layers: 
\begin{equation}
  \begin{aligned}
  &S_{C'}^{1} = \mathbf{NM}(S_{C'} + \mathbf{Att}(S_{C'}, S_{U_{n-1}}, S_{U_{n-1}})),\\
  &F_{C'} = \mathbf{NM}(S_{C'}^{1} + \mathbf{FF}(S_{C'}^{1})),
  \end{aligned}
\end{equation}
where $F_{C'}\in \mathbf{R}^{q_{C'} \times f_2}$, $f_2$ is the feature dimension of $F_{C'}$; $\mathbf{Att}$, $\mathbf{FF}$ and $\mathbf{NM}$ denote the multi-head attention layer, feed-forward layer and layer normalization, respectively; we input $S_{C'}$ as $Q$ of $\mathbf{Att}$, and take $S_{U_{n-1}}$ as $K$ and $V$ of $\mathbf{Att}$; the purpose of multi-head attention is to query the intention of the $\left(n-1\right)$-th utterance $U_{n-1}$ according to the semantic information of the $n-1$ past utterances $C'$; $S_{C'}^{1}$ contains both the contextual semantic feature of $C'$ and the intention information of $U_{n-1}$. Referring to KEMP, the emotion signal representation $e_{C'}$ is calculated as follows:
\begin{equation}
  e_{C'} = \mathbf{softmax}(\eta(C'))^\top \times S_{C'},
\end{equation}
where ${e_{C'}}^\top \in \mathbb{R}^{f_1}$; $S_{C'} \in \mathbb{R}^{q_{C'} \times f_1}$ denotes the semantic feature of the $n-1$ past utterances $C'$, and $\eta\left(C'\right) \in \mathbb{R}^{q_{C'}}$ is emotion intensity values provided by NRC\_VAD. Finally, we use $\mathbf{Dec}$ to generate the probability distribution of the virtual last utterance:
\begin{equation}
  P_{n} = \mathbf{Dec}(e_{C'}, F_{C'}, E_w(U_n)),
\end{equation}
where $\mathbf{Dec}$ is the Emotion-dependency decoder, and $E_w(U_n)$ is the word embedding of the real last utterance $U_n$; $E_w(U_n) \in \mathbb{R}^{m_n \times f}$, $m_n$ is the number of words in the real last utterance $U_n$. 

During training, the prediction loss is quantified by the negative log-likelihood with respect to the gold prediction $U_n$:
\begin{equation}
 \mathcal{L}^p = -\sum_{j = 1}^{m_n}\mathbf{log} P_{n}(U_n^p[j] = U_n[j] \vert (U_n[1], \cdots, U_n[j-1])),  
\end{equation}
where $U_n[j]$ denotes the $j$-th ground-truth word in $U_n$, and $U_n^p[j]$ denotes the $j$-th predicted word in the virtual last utterance $U_n^p$. In order to obtain the virtual last utterance $U_{n}^{p}$, we calculate the index corresponding to the maximum probability distribution of each element in $P_{n}$:
\begin{equation}
  IDX(U_n^p[j]) = \mathbf{argmax}(P_{n}(U_n^{p'}[j])),
\end{equation}
where $IDX(U_n^p[j])$ denotes the predicted index of the $j$-th word in $U_n^p$, and $P_{n}(U_n^{p'}[j])$ is the probability distribution of the $j$-th element in $P_{n}$. Then, the corresponding word is looked up in the dictionary based on the index.

\subsection{Attention based Intention Perception}\label{sec:percetion}
In this section, MAIFNet is exploited to integrate the dialogue history and last utterance to capture the speaker's intention information. We provide the entire dialogue $C$ and last utterance $U_n$ with emotion related concepts and extract their semantic features by Emotion context encoder:
\begin{equation}
  \begin{aligned}
  &S_{C} = \mathbf{EcEnc}_3(E_{C}), \\
  &S_{U_{n}} = \mathbf{EcEnc}_4(E_{U_{n}}),
  \end{aligned}
\end{equation}
where $E_{C} \in \mathbb{R}^{q_{C} \times f}$, $q_{C}$ is the number of words in $C$, and $q_{C}=m_1+\cdots+m_{n}$; $E_{U_{n}} \in \mathbb{R}^{m_n \times f}$. 

We believe that the last utterance contains the speaker's intention, so we take $S_{C}$ and $S_{U_{n}}$ together as the input to MAIFNet. Before that, we hope to increase the diversity of intention of the last utterance. First, we leverage the Transformer encoder to extract the semantic feature of the virtual last utterance $U_n^p$. Here, we denote the semantic feature of $U_n^p$ as $S_{U_{n}^p}$. Then, we concatenate $S_{U_{n}^p}$ with $S_{U_{n}}$ to obtain a new semantic feature representation $S_{U_{n}^{pr}}$. 

In the most important step, we leverage $\mathbf{MAIFNet}$ to fuse the semantic information of the dialogue history and that of the last utterance in order to perceive the speaker's intention. The process can be formulated as follows:
\begin{equation}
  \begin{aligned}
  &S_{C}^{1} = \mathbf{NM}(S_{C} + \mathbf{Att}(S_{C}, S_{U_n^{pr}}, S_{U_n^{pr}})),\\
  &F_{C} = \mathbf{NM}(S_{C}^{1} + \mathbf{FF}(S_{C}^{1})),
  \end{aligned}
\end{equation}
where $F_{C}\in \mathbf{R}^{q_{C} \times f_3}$; we input $S_{C}$ as $Q$ of $\mathbf{Att}$, and take $S_{U_n^p}$ as $K$ and $V$ of $\mathbf{Att}$; $S_{C}^{1}$ contains both the semantic information of the dialogue history and the intention information of the last utterance.

\subsection{Empathetic Response Generation}\label{sec:response}
Like the \nameref{sec:last} section, we calculate the emotional signal representation $e_C$ of the entire dialogue as follows:
\begin{equation}
e_C = \mathbf{softmax}(\eta(C))^\top \times S_C,
\end{equation}
where ${e_{C}}^\top \in \mathbb{R}^{f_4}$, $S_{C} \in \mathbb{R}^{q_{C} \times f_4}$, and $\eta\left(C\right) \in \mathbb{R}^{q_{C}}$. Finally, we generate the probability distribution of the empathetic response by employing the Emotion-dependency decoder:
\begin{equation}
  P_{R} = \mathbf{Dec}(e_{C}, F_{C}, E_w(R_{g})),
\end{equation}
where $E_w(R_{g})$ is the word embedding of the gold response. During training, we use the negative log-likelihood loss to optimize the response generation:
\begin{equation}
 \mathcal{L}^{r} = -\sum_{j = 1}^{t}\mathbf{log} P_{R}(R[j] = R_{g}[j] \vert (R_{g}[1], \cdots , R_{g}[j-1])),  
\end{equation}
where $t$ is the number of words in the gold response; $R_{g}[j]$ denotes the $j$-th word in the gold response, and $R[j]$ denotes the $j$-th word in the generated response.

To improve our model's emotion perception ability, we project the emotion signal representation $e_C$ into a vector whose length is same as the number of emotion categories and use the negative log-likelihood loss to optimize emotion perception learning:
\begin{equation}
\begin{aligned}
 &\widetilde{e}_C = \mathrm{W}_e\times {e_C}^\top,\\
 &P_e(\widetilde{e}_C) = \mathbf{softmax}(\widetilde{e}_C),\\
 &\mathcal{L}^{e} = -\mathbf{log}P_e(\widetilde{e}_{C}= e^\ast),
\end{aligned}
\end{equation}
where $\mathrm{W}_e \in \mathbb{R}^{q\times f_4}$, $q$ is the number of emotion categories, and $f_4$ is the dimension of the emotion signal representation. $e^\ast \in \mathbb{R}^q$ is the emotion label of the entire dialogue. As in KEMP, we adopt the emotional attention loss $\mathcal{L}^{a}$ to enforce the decoder to attach more attention to words with higher emotion intensity values:
\begin{equation}
  \mathcal{L}^{a} = \frac{1}{e}\sum_{i= 1}^{e} (\eta (C[i]) - a_i)^2,
\end{equation}
where $e$ is the number of words in the entire dialogue history, and $\eta (C[i])$ is the emotional intensity value of the $i$-th word in $C$; $a_i$ is the average attention score of $C[i]$, i.e., $a_i = \sum_{n = 1}^{H}a^n(R[j-1], C[i]) /H$. Here, $H$ is the number of attention heads, $a^n(R[j-1], C[i])$ is the attention score between the $(j-1)$-th generated word $R[j-1]$ and the $i$-th word $C[i]$ of the dialogue history $C$.

\subsection{Multi-task Training Objective}
We optimize all the parameters by minimizing the weighted summation of the above-mentioned four losses:
\begin{equation}
 \mathcal{L} = \mathcal{L}^{r} + \alpha_1\mathcal{L}^{p} + \alpha_2\mathcal{L}^{e} + \alpha_3\mathcal{L}^{a},
\end{equation}
where $\alpha_1$, $\alpha_2$, $\alpha_3$ are trade-off parameters. To better balance the optimizing rate of the parameters of prediction module and response module, we set $\alpha_1$ in this way: $\alpha_1 = \alpha'_1$ when $\mathcal{L}^{p} > \mathcal{L}^{r}$, $\alpha _1 = \alpha''_1$ when $\mathcal{L}^{p}\le \mathcal{L}^{r}$. Here, $\alpha'_1$ is a relatively larger value while $\alpha''_1$ is a relatively smaller value. This strategy can ensure that parameters of the \nameref{sec:last} section are optimized more quickly and get a more appropriate predicted utterance at the early training stage. On the contrary, at the late training stage, the parameters of the \nameref{sec:percetion} section and \nameref{sec:response} section are mainly optimized.

\section{Experiment}
\subsection{Dataset and Experimental Setup}
We evaluate the model on the EmpatheticDialogues~\cite{rashkin2018towards} dataset, an empathetic dialogue dataset containing $24850$ one-to-one open-domain dialogues, where the listener responds emphatically to the last utterance of the speaker. We use $17802$/$2628$/$2494$ dialogues as training/validation/testing set. The word embedding initialized by using the $300$-dimensional pre-trained Glove vectors~\cite{pennington2014glove}. The hidden dimension of each component of InferEM is set to $300$. The emotion lexical knowledge and commonsense knowledge are obtained form NRC\_VAD and ConceptNet in the same way as KEMP~\cite{li2022knowledge}. During testing the maximum of decoding steps is set as 30. The trade-off parameters $\alpha'_1$, $\alpha''_1$, $\alpha_2$, and $\alpha_3$ are set as $1.5$, $0.3$, $1.2$, and $0.12$, respectively. We train the model using Adam~\cite{kingma2014adam} with $\beta_1  = 0.9$ and $\beta_2  = 0.98$. The initial learning rate is set to $0.0001$ and varied during training in accordance with~\citeA{vaswani2017attention}, and the mini-batch size is set to $16$. All experiments are implemented in PyTorch on a single NVIDIA GeForce RTX 3090 with early stopping applied.

\subsection{Baselines and Evaluation Metrics}
\textbf{MoEL}~\cite{lin2019moel} is a Transformer-based model which designs different decoders to focus on different type of emotions. \textbf{MIME}~\cite{majumder2020mime} is a Transformer-based model which separates all emotions into positivity and negativity groups and decodes mimicking as well as non-mimicking response representations. \textbf{KEMP}~\cite{li2022knowledge} is a GAT and Transformer based model which uses commonsense knowledge to enrich dialogue history. \textbf{CEM}~\cite{sabour2022cem} is a Transformer-based model which focus on both affection and cognition aspects of empathy. To evaluate the effectiveness of the utterance prediction and intention perception components in our model, we design ablation studies as follows. \textbf{w/o SIP}: the model without the speaker's intention perception. \textbf{w/o LUP}: the model without the last utterance prediction.

\textbf{Automatic Evaluations}: With reference to KEMP, we adopt \textit{Emotion Accuracy}, \textit{Perplexity}~\cite{serban2015hierarchical}, \textit{Distinct-1}, and \textit{Distinct-2}~\cite{li2015diversity} to evaluate the performance of our model. \textbf{Human Evaluations}: following CEM, we carry out an aspect-based pairwise response comparison. That is, given 100 randomly sampled dialogue inputs, we ask $3$ annotators to compare responses from InferEM and baselines, and choose the better one on the following three aspects, including \textit{Empathy} (i.e., which response expresses more appropriate emotions), \textit{Relevance} (i.e., which response is more relevant with the given dialogue history) and \textit{Fluency} (i.e., which response has more readability and grammatical correctness). When they think two responses have much similar qualities and can't determine which one is better, they will give a judgment of \textit{Tie}.

\subsection{Results and Analysis}
\subsubsection{Comparison on Automatic Evaluation}
Table~\ref{tab:auto} shows that all automatic evaluation metrics of InferEM outperform the baseline KEMP by an obvious margin, which verifies the effectiveness of the last utterance prediction and intention perception components of our model to improve emotion perception, empathy expression and response diversity. Our model achieves the State-Of-The-Art performance in terms of Emotion Accuracy and Perplexity. The CEM performs better than our model in terms of Distinct-1 and Distinct-2 because it uses an additional Frequency-Aware Cross-Entropy loss~\cite{jiang2019improving} to penalize frequently occurring tokens. As InferEM obtains external knowledge in the same way as KEMP and is mainly compared with KEMP, we don't consider this loss. Therefore, we can compare InferEM with KEMP more fairly and verify the effectiveness of our proposed modules more convincingly.
\begin{table}[htbp] 
  \centering %
  \renewcommand{\arraystretch}{1.0}
  \setlength{\tabcolsep}{2pt}
  \caption{Comparison on automatic evaluation.}
  \begin{tabular}{c|c|c|c|c} \hline %
  \textbf{Models} & \textbf{Accuracy}(\%) & \textbf{Perplexity} & $\textbf{Distinct-1}$ & \textbf{Distinct-2}  \\ \hline
  MoEL & 32.00 &38.04 & 0.44 & 2.10  \\
  MIME &	34.24 &	37.09 &	0.47 & 1.91\\ 
  KEMP & 39.31 & 36.89 & 0.55 &	2.29\\ 
  CEM & 39.11 & 36.11 & \textbf{0.66} &	\textbf{2.99} \\
  InferEM & \textbf{39.98} & \textbf{31.26} & 0.59 & 2.60 \\ \hline
  w/o SIP & 39.14 &	31.43 &	0.52 & 2.08 \\ 
  w/o LUP& 38.15 & 33.73 & 0.52 & 2.09 \\ \hline
  \end{tabular}
  \label{tab:auto}
\end{table}

Table~\ref{tab:auto} shows that when the last utterance prediction or the intention perception module is removed, all the automatic evaluation metrics of our model become obviously worse, which means that the model's abilities to perceive intention and understand dialogue context decrease to some extent. This indicates that our proposed modules are beneficial for emotion perception and empathetic expression.

\subsubsection{Comparison on Human Evaluation}
\begin{table}[htbp] 
  \centering
  \renewcommand{\arraystretch}{1.0}
  \setlength{\tabcolsep}{3pt}
  \caption{Comparison on human evaluation.}
  \begin{tabular}{c|c|c|c|c} \hline 
  \textbf{Comparisons} & \textbf{Aspects} & \textbf{Win}(\%) & \textbf{Loss}(\%) & \textbf{Tie}(\%)  \\ \hline
   & Emp. &42.3 & 23.7 & 34.0  \\
   InferEM vs MoEL &	Rel. &	45.0 &	22.3& 32.7\\ 
   & Flu. & 24.7 & 19.3 &	56.0\\ \hline
   & Emp. &41.3 & 26.7 & 33.0 \\
   InferEM vs MiME &	Rel. &	42.3 &	23.7 & 34.0\\ 
   & Flu. & 22.3 & 17.7 &	60.0\\ \hline
   & Emp. &43.3 & 23.0 & 33.7  \\
   InferEM vs KEMP &	Rel. &	44.3 &	20.7 & 35.0\\ 
   & Flu. & 25.0 & 19.7 &	62.3\\ \hline
   & Emp. &32.7 & 23.3 & 44.0  \\
   InferEM vs CEM &	Rel. &	42.7 &	29.0 & 28.3\\ 
   & Flu. & 24.0 & 20.7 &	55.3\\ \hline
  \end{tabular}
  \label{tab:human}
\end{table}
Table~\ref{tab:human} shows that our model performs best on Empathy and Relevance. This suggests that our proposed modules can promote successful communication and generate more appropriate responses. As Transformer-based models can already generate fluent responses, our model is not noticeably different with baselines on Fluency.

\subsection{Case Studies}
Table~\ref{tab:comparision_case} shows two generated responses of our model and baselines. In the first case, InferEM outputs a suitable prediction of the last utterance $U_3^p$ with context-related words ``try my best", and generates a proper response with the excited emotion by replying with ``get it”. In the second case, InferEM generates the most appropriate response, containing the context-related words “help them up” and emotion-rated word “great”. These cases show that the last utterance prediction and intention perception modules of InferEM can improve mutual understanding and empathetic expression.
\begin{table}[htbp]
\centering %
\renewcommand{\arraystretch}{1.1}
\setlength{\tabcolsep}{4pt}
\caption{Generated responses from InferEM and baselines.}
\begin{tabular}{c|l} \toprule[0.5pt]
\textbf{Emotion} & \textbf{Excited} \\ 
\multirow{4}{*}{\textbf{History}} & $U_1$: I have a huge chance to win a ps4! \\
& $U_2$: Wow, please make it real.\\
& $U_3$: \multirow{2}{*}{\makecell[l]{There is only 2 other people in the final\\ round! winning a ps4 will be awesome!}}\\ 
& \\ \hline
\textbf{Gold} & Wow, i belief u can make it. \\
MoEL& I am sure you will have a great time! \\  
MIME& That is a great attitude to have!  \\ 
KEMP& \multirow{2}{*}{\makecell[l]{I agree with you. I have a huge fan of a few\\ years ago.}}\\
  & \\ 
CEM& I am sure you will do great!  \\ \hline
$U_3^p$& I will \underline{try my best} for it! \\ 
InferEM& That is great! I am sure you will \underline{get it}! \\
\bottomrule[0.2pt]
\toprule[0.2pt]
\textbf{Emotion} & \textbf{Proud} \\ 
\multirow{5}{*}{\textbf{History}} & $U_{1}$: \multirow{2}{*}{\makecell[l]{I am so happy for my son, went to his \\graduation from harvard last week.}}\\ 
& \\
& $U_{2}$: Wao, thats great, congrats!\\
& $U_{3}$: \multirow{2}{*}{\makecell[l]{Thank you! looks like there is going to be\\ another doctor in the family!}}\\ 
& \\ \hline
\textbf{Gold} & Thats good to know, keep it up. \\
MoEL& I am so happy for you! \\  
MIME& I am sure he will be fine.  \\ 
KEMP& I am sure you will do great! \\ 
CEM& That is great! I hope he's doing well  \\ \hline
$U_3^p$& I am so happy for him! \\ 
\multirow{2}{*}{InferEM} & \multirow{2}{*}{\makecell[l]{That is \underline{great}! I am sure you will be able to \underline{help}\\ \underline{them up}!}}\\ 
& \\
\bottomrule[0.5pt]
\end{tabular}
\label{tab:comparision_case}
\end{table}

Table~\ref{ablation_case} is a case where InferEM generates a different response when we remove the last utterance prediction module from it. Considering the impact of the predicted utterance $U_3^p$, InferEM generates a sounder response containing the context related and emotionally reasonable words ``cooperate better". It is obvious that if only the dialogue history is considered, the model generates a relatively monotonous response.
\begin{table}[htbp]
	\centering %
	\renewcommand{\arraystretch}{1.1}
	\setlength{\tabcolsep}{6pt}
	\caption{Generated responses from InferEM with and without the last utterance prediction module.}
	\begin{tabular}{c|l} 
	\toprule[0.5pt]
	\textbf{Emotion} & \textbf{Impressed} \\ 
	\multirow{5}{*}{\textbf{History}} & $U_1$: \multirow{3}{*}{\makecell[l]{I admire my boss because he does not act\\ like a boss. He acts more like a friend a-\\nd he is so down to earth.}} \\
	\\ \\
	& $U_2$: That is nice to have a friendly boss.\\
	& $U_3$: \multirow{2}{*}{\makecell[l]{He is not just friendly, he is a good lead-\\er and knows how to treat his workers.}}\\ 
	\\ \hline
	\textbf{Gold} & I envy you. That is cool.\\ \hline
	$U_3^p$& \multirow{2}{*}{\makecell[l]{Sure. It is my luck to work together with hi-\\m.}} \\ 
  \\
	InferEM& \multirow{2}{*}{\makecell[l]{That is good to hear. I believe that you will\\ \underline{cooperate} \underline{better} in the future.}}\\ 
	\\
	w/o LUP& I am glad you have a good leader. \\
	\bottomrule[0.5pt]
	\end{tabular}
	\label{ablation_case}
\end{table}

\section{Limitation}
Compared with recent large language models such as ChatGPT, InferEM may output flat responses such as ``I am sorry to hear that" in some cases. InferEM remains to be combined with pre-trained large language models to overcome this deficiency and generate more meaningful responses. It has to be mentioned that even though large language models show powerful performance, they are computationally very expensive and also require some specific components to further boost performance and enhance their interpretability. Therefore, the shallow frameworks like InferEM still deserve to be studied. 

Moreover, since the predicted last utterance and the real last utterance have different importance for generating responses, we believe that a proper attention mechanism designed to provide them with different weights will lead to better experimental results.

\section{Conclusion and Future Work}
Based on two key ideas of utterance prediction and intention perception, we propose a novel empathetic dialogue generation model named InferEM. The proposed intention fusion module is conducive to understanding the speaker's intention to generate more targeted responses. The predictive function of InferEM is beneficial for enhancing mutual understanding and generating more diverse responses. Furthermore, we design a multi-task learning strategy to better optimize the parameters of InferEM. Experimental results of automatic and human evaluations indicate that our proposed approach is effective in empathetic expression. In the future work, we will try utilizing superior information fusion methods to perceive intention.

\section*{Acknowledgements}
This work was supported in part by the National Natural Science Foundation of China under Grant 62236005, 61936004 and U1913602. We would like to thank the anonymous reviewers for their helpful remarks.

\balance
\bibliographystyle{apacite}
\setlength{\bibleftmargin}{.125in}
\setlength{\bibindent}{-\bibleftmargin}
\bibliography{inferem}

\end{document}